%% file: main.tex
\documentclass[letterpaper,10pt,conference]{ieeeconf}
\IEEEoverridecommandlockouts
\usepackage{graphicx}
\usepackage{capt-of}
\usepackage{amsmath,amssymb}
\usepackage[T1]{fontenc}
\usepackage{microtype}
\usepackage{booktabs,multirow,tabularx}
\usepackage{cite}
\usepackage{nicematrix}
\usepackage{url}
\usepackage[table]{xcolor}
\usepackage{xspace}
\usepackage{flushend}
\usepackage{placeins}
\usepackage{needspace}
\usepackage[normalem]{ulem}
\input{commands}
\title{\LARGE \bf
AVERT-VLN: Abstention-aware Visual Error Recovery and Training for Vision-and-Language Navigation}
\author{
Minrui Liu$^{1,\dagger}$, Jingke Wang$^{1,\dagger}$, Yuehao Huang$^{1}$, Hao Su$^{1}$, Jiajun L\"u$^{1}$, Yukai Ma$^{1,\ddagger}$, Yong Liu$^{1,*}$%
\thanks{$^{1}$ Institute of Cyber-Systems and Control, Zhejiang University, Hangzhou, China.}%
\thanks{$^{\dagger}$ These authors contributed equally to this work.}%
\thanks{$^{\ddagger}$ Project Lead.}%
\thanks{\raggedright $^*$ The corresponding author.}
}
\begin{document}
\bstctlcite{IEEEexample:BSTcontrol}
\maketitle
\thispagestyle{empty}
\pagestyle{empty}
\input{sections/00_abstract}
\input{sections/01_introduction}
\input{sections/02_related_work}
\input{sections/03_method}

\input{sections/04_experiments}

\Needspace{6\baselineskip}
\input{sections/05_limitations_conclusion}
\bibliographystyle{IEEEtran}
\bibliography{references}
\end{document}

%% file: commands.tex
\newcommand{\method}[1][ ]{Abstention-aware Visual Error Recovery and Training for Vision-and-Language Navigation\xspace}
\newcommand{\methodabbr}{AVERT-VLN\xspace}
\newcommand{\monitor}{Monitor\xspace}
\newcommand{\dataset}{\textsc{LostNav Dataset}\xspace}
\newcommand{\benchmark}{\textsc{LostAware Benchmark}\xspace}
\newcommand{\loststate}{\textsc{Lost}\xspace}
\newcommand{\passstate}{\textsc{Pass}\xspace}
\newcommand{\statepair}{\passstate{}/\loststate{}\xspace}

\definecolor{uvmblue}{HTML}{0072B2}
\definecolor{uvmorange}{HTML}{D55E00}
\definecolor{uvmgreen}{HTML}{009E73}
\definecolor{uvmyellow}{HTML}{E69F00}
\definecolor{uvmgray}{HTML}{6B7280}
\definecolor{uvmlightblue}{HTML}{E6F2F8}
\definecolor{uvmlightorange}{HTML}{FBEDE7}
\definecolor{uvmlightgray}{HTML}{F3F4F6}

\definecolor{tablegray}{HTML}{F2F2F2}
\definecolor{tablegray}{HTML}{F2F2F2}

%% file: sections/00_abstract.tex
\begin{abstract}
Deploying vision-and-language navigation (VLN) agents in unseen environments remains challenging because unfamiliar layouts and visual conditions can cause autonomous execution to go off track.
Rather than relying on continuous human supervision, a practical strategy is to let the agent selectively request corrective guidance when needed, use it to recover the ongoing task, and reuse the resulting corrective interactions to improve subsequent navigation.
We propose \method (\methodabbr), a closed-loop framework that uses a plug-in vision-language \monitor for online human-assisted recovery and offline preference learning.
The \monitor operates separately from navigation decision generation and assesses instruction--execution consistency from the instruction, visual history, and current observation.
To train the \monitor for deviation recognition, we construct \dataset with 20K counterfactual risk trajectories and rule-based deviation labels.
The \monitor is first fine-tuned on 40K normal trajectories to assess instruction progress and then jointly fine-tuned on normal and risk trajectories to recognize semantic deviations.
At runtime, Asynchronous Sidecar Monitoring evaluates execution alongside the navigation model.
When the controller accepts a \loststate verdict, it suspends autonomous execution and requests human guidance for recovery.
For offline policy improvement, Trajectory-Anchored Preference Learning converts deviation-associated failures into decision-level preference pairs under shared decision contexts, restricting supervision to the decisions targeted for correction.
Under human-assisted evaluation, the full \methodabbr system achieves success rates of 76.2\% and 66.3\% on the val-unseen splits of R2R-CE and RxR-CE, respectively.
The same monitoring and human-assisted recovery interface also improves success rates across the three evaluated navigation architectures.
More visualizations are available on our project page: \url{https://minrui-liu.github.io/avert-vln/}.
\end{abstract}

%% file: sections/01_introduction.tex
\section{Introduction}
\label{sec:introduction}
Vision-and-language navigation (VLN) enables embodied agents to follow natural-language route instructions using visual observations, supporting tasks such as delivery and guidance~\cite{anderson2018vision,krantz2020beyond}.
Recent navigation systems leverage vision-language models (VLMs) to interpret instructions, ground navigation targets, and generate navigation decisions~\cite{zheng2024navillm,zhang2024navid,wei2026ground,huang2025cogddn,huang2026wnm,li2026gn0,su2026sage}.
Nevertheless, deployment in unseen environments exposes VLN agents to layouts and visual conditions not covered during training, where execution may drift from the instruction and continued autonomous actions can compound such errors~\cite{hong2025general,ko2026active,yu2026user}.
A practical deployment system should therefore detect when execution goes
off track, request lightweight human correction when needed rather than rely
on continuous supervision, and reuse deployment failures and corrections to
improve subsequent navigation.

\input{figures/overview}

Prior work improves deployed navigation models through environment-specific adaptation and feedback-driven learning.
GSA-VLN~\cite{hong2025general} studies continual scene adaptation through repeated navigation in persistent environments. 
ATENA~\cite{ko2026active} uses episodic feedback for active test-time adaptation, while user-feedback-driven adaptation~\cite{yu2026user} incorporates episode-level success confirmations and goal-level corrections into model updates. 
Policy improvement from deployment experience is related to, but distinct from, deciding when to request assistance during an ongoing episode.
Just Ask~\cite{chi2020justask} enables agents to request assistance through either a confidence-based trigger or an explicit \textsc{Ask} action learned through reinforcement learning, and reuses human--agent interaction histories for further policy learning.
These lines of work motivate a complementary question: \textit{how can a VLN agent detect when its execution goes off track, seek human assistance to recover the ongoing task, and turn deployment failures into supervision for improving navigation in unseen environments?}

To address this question, we propose \textbf{A}bstention-aware \textbf{V}isual \textbf{E}rror \textbf{R}ecovery and \textbf{T}raining for \textbf{V}ision-and-\textbf{L}anguage \textbf{N}avigation \textbf{(AVERT-VLN)}, a closed-loop framework for continuous VLN deployment that connects off-track detection, human-assisted recovery, and failure-driven policy improvement through a shared vision-language \monitor, as illustrated in Fig.~\ref{fig:overview}.
The \monitor assesses instruction--execution consistency outside the navigation model's action-generation process, using the instruction, visual history, and current observation.
To train this assessor, we construct \dataset with 20K counterfactual risk trajectories and rule-based deviation labels.
Starting from Qwen3.5-4B~\cite{qwen2026qwen35_4b}, we first fine-tune the \monitor on 40K normal trajectories to assess instruction progress, then jointly on normal and risk trajectories to recognize semantic deviations.
At runtime, Asynchronous Sidecar Monitoring evaluates execution alongside the navigation model, allowing navigation to proceed while monitoring requests are pending.
When the controller accepts a \loststate verdict, it suspends autonomous execution and requests a pixel goal in the current view or a local turning command from a human operator.
After the corrective action is executed and a new observation becomes available, autonomous navigation and monitoring resume.

For offline policy improvement, a deviation verdict alone does not specify which decision to revise or which alternative to prefer.
We therefore develop Trajectory-Anchored Preference Learning, which uses \monitor judgments to locate candidate decision anchors associated with observed deviations.
At each anchor, corrective feedback obtained during human-assisted recovery
is used to construct a task-consistent preferred decision, which is paired with the original model output under the same decision context comprising the instruction, visual history, and current observation.
We optimize the navigation model with a decision-focused objective based on direct preference optimization (DPO)~\cite{rafailov2023direct}, applying preference supervision only to the decision outputs targeted for correction.
Together, Asynchronous Sidecar Monitoring and Trajectory-Anchored Preference
Learning connect online recovery with subsequent policy improvement.

{\setlength{\parskip}{0pt}
\textbf{Our contributions are summarized as follows:}\par
\begin{itemize}[\setlength{\topsep}{2pt}]
    \item We introduce \methodabbr, a closed-loop framework for VLN deployment connecting off-track detection, human-assisted recovery, and failure-driven policy improvement through a shared vision-language \monitor.
    \item We develop Asynchronous Sidecar Monitoring for abstention and human-assisted recovery, and Trajectory-Anchored Preference Learning to turn deviation-associated failures and corrective feedback into decision-level preference pairs under shared decision contexts.
    \item We construct \dataset for \monitor training and \benchmark for evaluating semantic deviation recognition.
    Experiments on the val-unseen splits of R2R-CE and RxR-CE show human-assisted recovery gains across three navigation architectures, while autonomous ablations isolate the effect of failure-derived preference learning from online recovery.
\end{itemize}}

%% file: figures/overview.tex
\begin{figure}[t]
    \centering
    \includegraphics[width=\columnwidth]{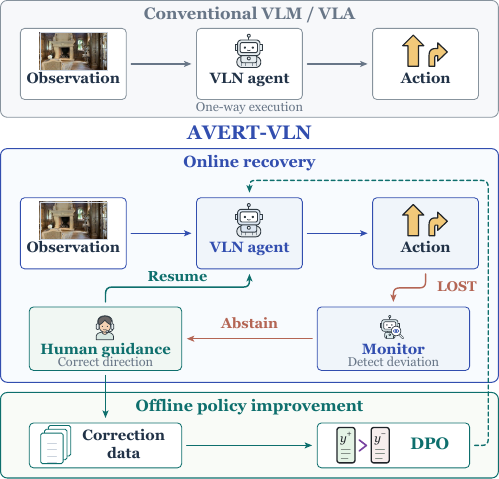}
    \caption{\textbf{Conceptual comparison of conventional VLN and \methodabbr.} \methodabbr forms a closed deployment loop in which the \monitor detects off-track execution, human guidance supports recovery when needed, and the resulting corrections are reused for offline policy improvement.}
    \label{fig:overview}
    \vspace{-20pt}
\end{figure}

%% file: sections/02_related_work.tex
\section{Related Work}
\label{sec:related_work}

\subsection{Execution Assessment and Interactive Assistance}
Prior VLN work has explored execution assessment and interactive assistance for handling ambiguous or off-track states.
VNLA~\cite{nguyen2019vision} enables an agent to query an advisor for language subgoals, whereas Just Ask~\cite{chi2020justask} requests assistance through either an action-confidence threshold or a learned \textsc{Ask} action.
Alongside interactive assistance, progress estimation guides navigation~\cite{ma2019self}, while recent VLM-based methods reason about agent state or semantic progress to condition navigation decisions~\cite{guo2026awarevln,wang2026progressthink}.
Navigation Heads detects path deviations from internal attention heads in a frozen navigation model and invokes a lightweight rollback policy for recovery~\cite{jeong2026navigationheads}.
In contrast, our \monitor operates outside the navigation model's native action-generation process and provides a shared instruction--execution assessment used by Asynchronous Sidecar Monitoring for human-assisted recovery across navigation architectures.

\subsection{Learning from Navigation Failures}
Learning from navigation failures depends on where to anchor corrective supervision and which target to assign.
CorrectNav~\cite{yu2026correctnav} derives self-correction supervision for perception and action decisions from mined error trajectories, whereas BudVLN~\cite{he2026budvln} retrospectively reanchors corrective supervision to valid historical states.
SACA~\cite{li2026let} uses step-aware assessment to identify valid prefixes and divergence points, extracting dense corrective supervision from failed trajectories.
AeroDPO~\cite{xu2026aerodpo} uses simulator rollback and privileged interventions to construct collision-avoidance preference pairs.
In contrast, our Trajectory-Anchored Preference Learning uses \monitor judgments to locate deviation-associated decision anchors and corrective feedback to construct task-consistent preference pairs under shared decision contexts, while restricting preference optimization to the decision outputs targeted for correction.

%% file: sections/03_method.tex
\section{Method}
\label{sec:problem_framework}

\input{figures/pipeline}

\subsection{Overview}
\label{sec:framework_overview}

\methodabbr provides a closed-loop framework for VLN deployment in unseen environments.
A shared vision-language \monitor assesses instruction--execution
consistency while the navigation model retains its native action-generation process.
During execution, Asynchronous Sidecar Monitoring uses \monitor verdicts to coordinate abstention and human-assisted recovery.
For subsequent policy improvement, Trajectory-Anchored Preference Learning uses deviation-associated failures and corrective feedback to construct decision-level preference supervision under shared decision contexts.

\subsection{Counterfactual Risk-Trajectory Construction}
\label{sec:risk_data}
We define a navigation deviation as an observable inconsistency between the instruction and the agent's target selection or navigation behavior.
Geometric distance to a reference trajectory is insufficient for identifying such deviations: valid alternative routes may depart from the reference path, whereas an incorrect target may remain geometrically nearby.
Accordingly, we construct training data covering instruction-consistent and deviating executions.
We assign a \loststate label only when execution exhibits a verifiable conflict with the instruction, and pair it with a diagnostic description grounded in the instruction, visual history, and current observation.
As shown in Fig.~\ref{fig:lostnav_pipeline}, \dataset follows a three-step pipeline comprising intervention generation, candidate validation, and simulator rollout with annotation.

\subsubsection{Failure Case Generation}
We use the instance and region dictionaries from VL-LN Bench~\cite{huang2025vlln} to identify scene entities for region- and object-level interventions.
At a fixed execution state, we intervene on target selection and simulate the counterfactual continuation.
To generate controlled deviations at different semantic levels, we factor each intervention into a behavior mode \(b\in\mathcal{B}\) and a target granularity \(g\in\mathcal{G}\), where \(\mathcal{B}=\{\mathrm{explore},\mathrm{exploit}\}\) and \(\mathcal{G}=\{\mathrm{region},\mathrm{object}\}\).
The behavior mode controls target commitment: under \(\mathrm{explore}\), the intervention keeps the agent searching despite a visible subgoal, whereas under \(\mathrm{exploit}\), it redirects the agent toward an incorrect target.
The granularity specifies whether the intervention operates at the region or object level, yielding four intervention types in \(\mathcal{B}\times\mathcal{G}\).

\subsubsection{Candidate Validation}
Before simulator rollout, structural checks verify episode--instruction correspondence, action--observation alignment, temporal ordering, and valid simulator-state restoration.
Physical and semantic checks verify the selected entity is annotated, visible, navigable from the restored state, and consistent with the intended intervention.
Only candidates passing both checks are executed.

\subsubsection{Simulator Rollout and Annotation}
Executing the validated interventions in the simulator produces \(\mathcal{D}_{\mathrm{lost}}\), comprising 20K counterfactual risk trajectories with rendered observations.
Rule-based criteria assign deviation labels, while \textit{\mbox{doubao-seed-2-0-pro-260215}} generates the corresponding diagnostic descriptions.
Scene metadata and privileged navigation states are used only for data construction and verification and are never provided as \monitor inputs.

\subsection{\monitor Training}
\label{sec:monitor_training}

Assessing instruction--execution consistency requires tracking instruction progress over the visual history and interpreting the current observation in that context.
Let \(M_\phi\) denote the \monitor with parameters \(\phi\). We initialize the \monitor from Qwen3.5-4B~\cite{qwen2026qwen35_4b} and fine-tune it in two stages.
Both stages condition on the instruction \(\mathcal{I}\), a history \(\mathcal{H}_t\) of at most eight temporally ordered frames, and the current observation \(o_t\).
We structure the diagnostic description as
\begin{equation}
    \label{eq:monitor_reasoning}
    r_t
    =
    r_t^{\mathrm{hist}}
    \oplus
    r_t^{\mathrm{cur}},
\end{equation}
where \(r_t^{\mathrm{hist}}\) summarizes instruction progress supported by the visual history, \(r_t^{\mathrm{cur}}\) assesses the current observation against that progress, and \(\oplus\) denotes text concatenation.

\subsubsection{Stage 1: Normal-Trajectory Reasoning}
Using the same annotator as for the risk trajectories, we annotate 40K normal trajectories with diagnostic descriptions to form \(\mathcal{D}_{\mathrm{pass}}\).
We first fine-tune the \monitor on this dataset to model instruction progress and diagnose the current state under normal execution.

\subsubsection{Stage 2: Joint Pass/Lost Fine-Tuning}
Starting from the Stage-1 checkpoint, we jointly fine-tune the \monitor on \(\mathcal{D}_{\mathrm{pass}}\cup\mathcal{D}_{\mathrm{lost}}\).
Normal examples retain the same diagnostic supervision as in Stage 1, while risk examples introduce semantic-deviation diagnostics and the corresponding \loststate verdicts.
Joint training therefore exposes the \monitor to both instruction-consistent and deviating executions under a common input--output format.

At inference time, the \monitor generates a diagnostic response from which the controller obtains a \passstate or \loststate verdict, without a separate classification head.

\input{figures/dataset}

\subsection{Asynchronous Sidecar Monitoring}
\label{sec:online_recovery}
As shown in the upper module of Fig.~\ref{fig:online_recovery}, Asynchronous Sidecar Monitoring closes the online recovery loop by coupling \monitor-based deviation assessment with human intervention, without modifying the navigation model's native action-generation process.
The navigation model issues its native navigation commands, while the controller executes them and the \monitor asynchronously evaluates the resulting observations in micro-batches.
Given the instruction \(\mathcal{I}\), visual history \(\mathcal{H}_t\) of the active execution, and current observation \(o_t\), the \monitor generates
\begin{equation}
    \label{eq:sidecar_monitor_response}
    v_t=M_\phi(\mathcal{I},\mathcal{H}_t,o_t),
\end{equation}
where \(v_t\) is a diagnostic containing a deviation verdict \(z_t\in\{\text{\passstate},\text{\loststate}\}\) and supporting diagnostic text \(r_t\).

The commit frontier denotes the latest monitoring boundary such that all committed verdicts up to that boundary are \passstate.
Previously executed actions are not reverted; recovery starts from the current execution state.
Committing a \loststate verdict halts unexecuted actions and invalidates pending monitoring requests associated with the current execution, triggering human-assisted recovery.
A remote human supplies a pixel goal in the current view or a local turning command, which the controller converts into an executable local action.
Navigation and asynchronous monitoring resume once the corrective action yields a new observation.

\subsection{Trajectory-Anchored Preference Learning}
\label{sec:preference_alignment}
For offline policy improvement, Trajectory-Anchored Preference Learning converts deviation-associated failures and corrective feedback into decision-level preferences, as shown in the lower module of Fig.~\ref{fig:online_recovery}.
Because a \loststate verdict identifies a failure but not which decision to revise or which alternative to prefer, we use \monitor judgments to locate decision anchors and corrective feedback to construct task-consistent preference pairs under shared decision contexts.

\subsubsection{Anchored Preference Construction}
For each policy-generated trajectory receiving a \loststate verdict, we trace the detected deviation to the nearest preceding boundary judged \passstate and treat it as a decision anchor.
At the anchor context \(x\), comprising the instruction, visual history, and current observation, corrective feedback provides a task-consistent preferred decision \(y^{+}\), while the original output at the same decision stage serves as \(y^{-}\).
Specifically, a pixel-goal correction defines the preferred visual-grounding decision, whereas a local-turn correction defines the preferred direction-selection decision.
Each sample \(d=(x,y^+,y^-,k)\in\mathcal{D}_{\mathrm{pref}}\) additionally records the corrected stage \(k\), corresponding to direction selection or visual grounding.

\subsubsection{Decision-Focused Optimization}
We initialize the trainable policy \(\pi_\theta\) and frozen reference policy \(\pi_{\mathrm{ref}}\) from the same System~2 checkpoint of DualVLN~\cite{wei2026ground}.
For each sample \(d=(x,y^+,y^-,k)\), the stage \(k\) determines a candidate-specific mask \(m\) selecting only tokens from the decision stage targeted for correction; \(m\) is instantiated separately for each candidate and shared between \(\pi_\theta\) and \(\pi_{\mathrm{ref}}\) when scoring that candidate.
Both policies score each candidate \(y\) under the same context \(x\) and mask \(m\) using
\begin{equation}
    \label{eq:masked_decision_score}
    s_{\pi}(y\mid x;m)
    =
    \frac{\sum_j m_j\log\pi(y_j\mid x,y_{<j})}
    {\sum_j m_j},
\end{equation}
where \(m_j\in\{0,1\}\).

The reference-adjusted preference margin is
\begin{equation}
    \label{eq:preference_margin}
    \begin{aligned}
    \Delta_\theta(d)
    ={}&
    \bigl[
    s_{\pi_\theta}(y^+\mid x;m)
    -
    s_{\pi_\theta}(y^-\mid x;m)
    \bigr]
    \\
    &-
    \bigl[
    s_{\pi_{\mathrm{ref}}}(y^+\mid x;m)
    -
    s_{\pi_{\mathrm{ref}}}(y^-\mid x;m)
    \bigr].
    \end{aligned}
\end{equation}
We optimize a decision-focused objective based on DPO:
\begin{equation}
    \label{eq:policy_preference_objective}
    \begin{aligned}
    \mathcal{L}_{\mathrm{policy}}
    ={}&-\mathbb{E}_{d\sim\mathcal{D}_{\mathrm{pref}}}
    \Bigl[
    \log\sigma\!\left(\beta\Delta_\theta(d)\right)\\
    &\quad+\lambda s_{\pi_\theta}(y^+\mid x;m)
    \Bigr].
    \end{aligned}
\end{equation}
Here, \(\sigma\) is the sigmoid function, \(\beta>0\) scales the margin, and \(\lambda\geq0\) weights the preferred-output likelihood term.

%% file: figures/pipeline.tex
\begin{figure*}[t!]
    \centering
    \includegraphics[width=\textwidth]{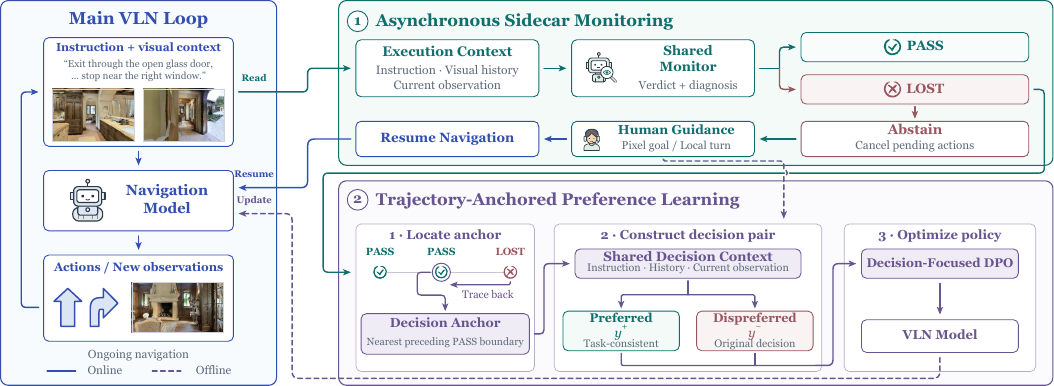}
    \caption{\textbf{Online recovery and offline preference learning in \methodabbr.}
    (1) Asynchronous Sidecar Monitoring assesses execution alongside the navigation model and triggers abstention and human-assisted recovery
    upon a \loststate verdict.
    (2) Trajectory-Anchored Preference Learning traces deviations to the nearest preceding \passstate boundary, pairs correction-derived preferred with the original output under a shared context, and updates the navigation model offline using decision-focused DPO.}
    \label{fig:online_recovery}
    \vspace{-8pt}
\end{figure*}

%% file: figures/dataset.tex
\begin{figure}[t!]
    \centering
    \includegraphics[width=\columnwidth]{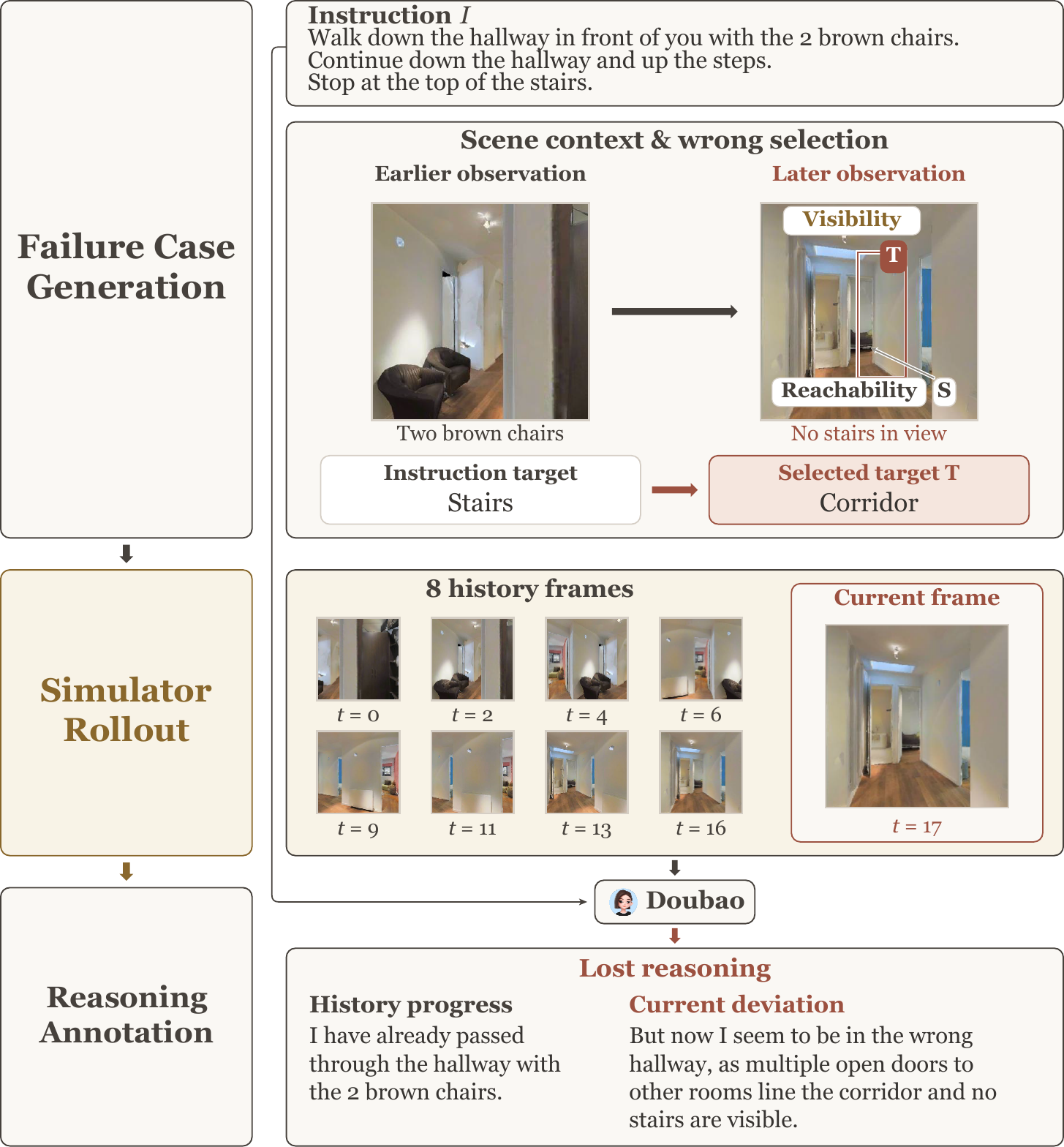}
    \caption{\textbf{Construction of \dataset.} Simulator rollouts of validated target-selection interventions yield counterfactual risk trajectories with rule-based deviation labels and diagnostic annotations.}
    \label{fig:lostnav_pipeline}
    \vspace{-10pt} 
\end{figure}

%% file: sections/04_experiments.tex
\section{Experiments}
\label{sec:experiments}

\newcommand{\cmark}{\ensuremath{\checkmark}}

\input{tables/ce}

\subsection{Experimental Setup}
\label{sec:experimental_setup}
We evaluate the \monitor's ability to recognize instruction deviations, the effect of preference learning on autonomous navigation, and the benefit of online monitoring with human-assisted recovery across navigation architectures.

\subsubsection{Navigation Benchmarks and Metrics}
We evaluate closed-loop navigation on Room-to-Room in Continuous Environments (R2R-CE) and Room-across-Room in Continuous Environments (RxR-CE), using their validation splits in unseen environments (val-unseen)~\cite{krantz2020beyond,ku2020rxr}.
We report navigation error (NE), oracle success (OS), success rate (SR), success weighted by path length (SPL)~\cite{anderson2018spl}, and normalized dynamic time warping (nDTW)~\cite{ilharco2019ndtw}.
NE is measured in meters; the remaining navigation scores are percentages.
Table~\ref{tab:full_navigation_comparison} also records the observation requirements of each method: panoramic observations (Pano.), odometry (Odo.), depth, and single-view RGB (S.RGB).

\subsubsection{\textup{\benchmark}}
We evaluate deviation recognition on paired \statepair samples from English RxR-CE~\cite{krantz2020beyond,ku2020rxr}.
The benchmark pairs failed Follower trajectories with Guide reference trajectories for the same scene, instruction, and goal.
Guides describe prescribed routes while traversing them, whereas Followers navigate independently from these instructions.
Within failed Follower trajectories, we identify candidate states using geometric deviation, failure to recover, inconsistency with subsequent Guide regions, and object-context mismatch.
States from temporally stable high-score segments are paired with Guide states at matched normalized trajectory progress.
Model-assisted review and human adjudication retain 1,095 pairs, each containing one \loststate Follower state and one \passstate Guide state.
All 2,190 samples provide the instruction, visual history, and current observation and are reserved for evaluation.
We report recall and F1 with \loststate as the positive class.

\subsection{Implementation Details}
\label{sec:implementation_details}

\input{figures/monitor-diagnostics-recognition}

\subsubsection{\monitor Training}
We initialize the \monitor from Qwen3.5-4B and perform full-parameter fine-tuning on eight NVIDIA H20 GPUs.
Following Section~\ref{sec:monitor_training}, Stage~1 trains on the normal-trajectory dataset \(\mathcal{D}_{\mathrm{pass}}\) for one epoch, and Stage~2 continues for one epoch on \(\mathcal{D}_{\mathrm{pass}}\cup\mathcal{D}_{\mathrm{lost}}\).

\subsubsection{Navigation Training}
For Trajectory-Anchored Preference Learning, we optimize System~2 of DualVLN.
We train it for two epochs using the objective in Eq.~\eqref{eq:policy_preference_objective}, with \(\beta=0.5\) and a preferred-output weight of \(\lambda=0.1\).

\subsubsection{Online Monitoring}
The navigation model and \monitor run through the Asynchronous Sidecar Monitoring interface.
Monitoring requests use a maximum batch size of four and a coalescing timeout of 0.05~s.
A committed \loststate verdict triggers human-assisted recovery using a pixel goal or a local turning command, as described in Section~\ref{sec:online_recovery}.

\subsection{Experimental Results}
\label{sec:experimental_results}
\subsubsection{Navigation Performance}
\label{sec:navigation_results}

Table~\ref{tab:full_navigation_comparison} evaluates \methodabbr across three navigation architectures on the R2R-CE and RxR-CE val-unseen splits.
The \(^{\dagger}\) configurations add the shared \monitor and human-assisted recovery to StreamVLN~\cite{wei2025streamvln}, InternVLA-N1 (System~2)~\cite{internrobotics2025internvla}, and DualVLN~\cite{wei2026ground} checkpoints without additional navigation-model training.
The final \(^{\ddagger}\) configuration uses the DualVLN dual-system architecture.
System~2 is optimized through Trajectory-Anchored Preference Learning, followed by fine-tuning of the corresponding System~1 to adapt low-level execution to the updated System~2.
The resulting model is evaluated with the same monitoring and human-assisted recovery.

The full configuration achieves SRs of 76.2\% on R2R-CE and 66.3\% on RxR-CE.
Compared with DualVLN using the same monitoring and recovery interface, SR increases by 0.7 points on both benchmarks, while SPL increases by 1.1 and 5.0 points, respectively.
The improvement is pronounced in success-weighted path efficiency, particularly on RxR-CE, alongside a modest increase in task completion.

Across the three released navigation architectures, online monitoring and human-assisted recovery increase SR and reduce NE on both benchmarks.
SR gains range from 6.8 to 14.5 percentage points on R2R-CE and from 2.1 to 8.3 points on RxR-CE, supporting the applicability of the shared interface across the evaluated architectures.
For InternVLA-N1 on RxR-CE, nDTW decreases from 62.6 to 57.3, indicating lower agreement with the reference trajectory.
Backtracking and corrective movements during recovery may contribute to this reduction in trajectory fidelity.

\input{figures/dpo_qualitative_cases}

\subsubsection{Lost-State Recognition and Diagnostic Reasoning}
We compare Monitor-4B with 17 open-source VLMs~\cite{bai2025qwen25vl,bai2025qwen3vl,qwen2026qwen35,qwen2026qwen36,zhu2025internvl3,wang2025internvl35}.
Monitor-4B achieves the highest recall and F1 at 63.38\% and 59.52\%, respectively, as shown in Fig.~\ref{fig:monitor_radar}(b).
These exceed its Qwen3.5-4B initialization by 34.43 and 21.35 percentage points, and Qwen3.5-27B, the strongest F1 baseline, by 20.37 and 9.73 points, respectively.
Its precision is 56.10\%, compared with 59.10\% for Qwen3.5-27B, reflecting a precision--recall trade-off.

We use GPT-5.6 Luna to score Monitor-4B and six baselines on the same samples using the same judging prompt.
The six-dimensional rubric evaluates consistency with prior observations (\textit{history consistency}), grounding in visible objects and spatial relations (\textit{spatial grounding}), identification of completed and remaining instruction steps (\textit{instruction progress}), explanation of instruction--execution conflicts (\textit{deviation diagnosis}), observational support for diagnostic claims (\textit{evidence faithfulness}), and the usefulness of the diagnosis for correcting execution (\textit{recovery usefulness}).
Scores are on a 0--100 scale.
Monitor-4B obtains the highest reported scores across all six dimensions in Fig.~\ref{fig:monitor_radar}(a).
The largest gaps occur in history consistency and recovery usefulness, where it scores 74.93 and 61.44, exceeding the strongest baseline in each dimension by 21.45 and 11.17 points, respectively.

\subsection{Component Ablation}
\label{sec:component_ablation}

\input{tables/ablation}

We use System~2 of DualVLN as the baseline for component ablations on RxR-CE val-unseen (Table~\ref{tab:component_ablation}).
Combining DPO with online monitoring and human-assisted recovery yields 60.21\% SR.
Removing online recovery or preference learning reduces SR by 5.18 or 2.43 points, respectively.

Additional SFT controls for the effect of extra training data.
With online monitoring and human assistance disabled, DPO outperforms additional SFT by 1.62 percentage points in SR, supporting failure-derived preference learning for autonomous navigation.
In the selected cases shown in Fig.~\ref{fig:dpo_qualitative_cases}, +DPO uses the specified landmarks to choose its direction and reaches the instructed destination, completing tasks that Original fails to finish.
Online monitoring with human-assisted recovery also improves task success for System~2 without preference training by enabling correction during execution.

%% file: tables/ce.tex
\definecolor{obsgray}{HTML}{EFEFEF}
\definecolor{streamcolor}{HTML}{F2F7FA}
\definecolor{interncolor}{HTML}{E1EDF5}
\definecolor{dualcolor}{HTML}{CDDFEC}

\begin{table*}[t]
    \caption{\textbf{Navigation Performance on R2R-CE and RxR-CE Val-Unseen Splits.}}
    \label{tab:full_navigation_comparison}
    \centering
    \footnotesize
    \setlength{\tabcolsep}{5.5pt}
    \renewcommand{\arraystretch}{1.13}

    \begin{NiceTabular}{lcccccccccccc}
        \CodeBefore
        \rowcolor{streamcolor}{22}
        \rowcolor{streamcolor}{25}
        \rowcolor{interncolor}{23}
        \rowcolor{interncolor}{26}
        \rowcolor{dualcolor}{24}
        \rowcolor{dualcolor}{27-28}
        \Body
        \toprule

        \multirow{2}{*}{\textbf{Method}}
        & \multicolumn{4}{c}{\textbf{Observation}}
        & \multicolumn{4}{c}{\textbf{R2R-CE Val-Unseen}}
        & \multicolumn{4}{c}{\textbf{RxR-CE Val-Unseen}} \\
        \cmidrule(lr){2-5}
        \cmidrule(lr){6-9}
        \cmidrule(lr){10-13}

        & \textbf{Pano.}
        & \textbf{Odo.}
        & \textbf{Depth}
        & \textbf{S.RGB}
        & \textbf{NE} $\downarrow$
        & \textbf{OS} $\uparrow$
        & \textbf{SR} $\uparrow$
        & \textbf{SPL} $\uparrow$
        & \textbf{NE} $\downarrow$
        & \textbf{SR} $\uparrow$
        & \textbf{SPL} $\uparrow$
        & \textbf{nDTW} $\uparrow$ \\

        \midrule
        HPN+DN$^{*}$~\cite{krantz2021waypoint}
        & \checkmark & \checkmark & \checkmark &
        & 6.31 & 40.0 & 36.0 & 34.0 & -- & -- & -- & -- \\

        CMA$^{*}$~\cite{hong2022bridging}
        & \checkmark & \checkmark & \checkmark &
        & 6.20 & 52.0 & 41.0 & 36.0 & 8.76 & 26.5 & 22.1 & 47.0 \\

        GridMM$^{*}$~\cite{wang2023gridmm}
        & \checkmark & \checkmark & \checkmark &
        & 5.11 & 61.0 & 49.0 & 41.0 & -- & -- & -- & -- \\

        ETPNav$^{*}$~\cite{an2024etpnav}
        & \checkmark & \checkmark & \checkmark &
        & 4.71 & 65.0 & 57.0 & 49.0 & 5.64 & 54.7 & 44.8 & 61.9 \\

        ScaleVLN$^{*}$~\cite{wang2023scaling}
        & \checkmark & \checkmark & \checkmark &
        & 4.80 & -- & 55.0 & 51.0 & -- & -- & -- & -- \\

        \midrule
        InstructNav~\cite{long2024instructnav}
        & \checkmark & \checkmark & \checkmark & \checkmark
        & 6.89 & -- & 31.0 & 24.0 & -- & -- & -- & -- \\

        R2R-CMTP~\cite{chen2021topological}
        & \checkmark & \checkmark & \checkmark &
        & 7.90 & 38.0 & 26.4 & 22.7 & -- & -- & -- & -- \\

        LAW~\cite{raychaudhuri2021language}
        & & \checkmark & \checkmark & \checkmark
        & 6.83 & 44.0 & 35.0 & 31.0 & 10.90 & 8.0 & 8.0 & 38.0 \\

        CM2~\cite{georgakis2022cross}
        & & \checkmark & \checkmark & \checkmark
        & 7.02 & 41.5 & 34.3 & 27.6 & -- & -- & -- & -- \\

        WS-MGMap~\cite{chen2022weakly}
        & & \checkmark & \checkmark & \checkmark
        & 6.28 & 47.6 & 38.9 & 34.3 & -- & -- & -- & -- \\

        ETPNav + FF~\cite{wang2024sim}
        & & \checkmark & \checkmark & \checkmark
        & 5.95 & 55.8 & 44.9 & 30.4 & 8.79 & 25.5 & 18.1 & -- \\

        Seq2Seq~\cite{krantz2020beyond}
        & & & \checkmark & \checkmark
        & 7.77 & 37.0 & 25.0 & 22.0 & 12.10 & 13.9 & 11.9 & 30.8 \\

        CMA~\cite{krantz2020beyond}
        & & & \checkmark & \checkmark
        & 7.37 & 40.0 & 32.0 & 30.0 & -- & -- & -- & -- \\

        \midrule
        NaVid~\cite{zhang2024navid}
        & & & & \checkmark
        & 5.47 & 49.1 & 37.4 & 35.9 & -- & -- & -- & -- \\

        MapNav~\cite{zhang2025mapnav}
        & & & & \checkmark
        & 4.93 & 53.0 & 39.7 & 37.2 & -- & -- & -- & -- \\

        NaVILA~\cite{cheng2024navila}
        & & & & \checkmark
        & 5.22 & 62.5 & 54.0 & 49.0 & 6.77 & 49.3 & 44.0 & 58.8 \\

        UniNaVid~\cite{zhang2024uni}
        & & & & \checkmark
        & 5.58 & 53.3 & 47.0 & 42.7 & 6.24 & 48.7 & 40.9 & -- \\

        BudVLN~\cite{he2026budvln}
        & & & & \checkmark
        & 4.74 & 65.6 & 57.6 & 51.1 & 5.79 & 56.1 & 46.6 & 63.2 \\

        SACA~\cite{li2026let}
        & & & & \checkmark
        & 4.57 & 64.9 & 60.3 & 55.1 & 4.90 & 60.3 & 49.8 & 62.1 \\

        \midrule
        StreamVLN~\cite{wei2025streamvln}
        & & & & \checkmark
        & 4.98 & 64.2 & 56.9 & 51.9 & 6.22 & 52.9 & 46.0 & 61.9 \\

        InternVLA-N1 (System~2)~\cite{internrobotics2025internvla}
        & & & & \checkmark
        & 4.89 & 60.6 & 55.4 & 52.1 & 6.41 & 49.5 & 41.8 & 62.6 \\

        DualVLN~\cite{wei2026ground}
        & & & & \checkmark
        & 4.05 & 70.7 & 64.3 & 58.5
        & 4.58 & 61.4 & \textbf{51.8} & 70.0 \\

        \midrule
        \textbf{\methodabbr}$^{\dagger}$ \textbf{(StreamVLN)}
        & & & & \checkmark
        & 3.31 & 82.3 & 63.7 & 42.6 & 5.58 & 55.0 & 47.0 & 63.7 \\

        \textbf{\methodabbr}$^{\dagger}$ \textbf{(InternVLA-N1)}
        & & & & \checkmark
        & 2.97 & 78.2 & 69.9 & 28.7 & 4.53 & 57.8 & 32.7 & 57.3 \\

        \textbf{\methodabbr}$^{\dagger}$ \textbf{(DualVLN)}
        & & & & \checkmark
        & 2.61 & 80.7 & 75.5 & 62.3 & 3.93 & 65.6 & 43.0 & 71.7 \\

        \textbf{\methodabbr (DPO + DualVLN)}$^{\ddagger}$
        & & & & \checkmark
        & \textbf{2.60} & \textbf{82.5} & \textbf{76.2} & \textbf{63.4}
        & \textbf{3.83} & \textbf{66.3} & 48.0 & \textbf{72.0} \\

        \bottomrule
    \end{NiceTabular}

    \par\vspace{2pt}
    \begin{minipage}{\textwidth}
        \footnotesize
        $^{*}$ Methods using pretrained waypoint predictors.
        $^{\dagger}$ \monitor and human-assisted recovery with released navigation
        checkpoints and no additional navigation-model training.
        $^{\ddagger}$ DPO-trained System~2 with subsequently fine-tuned System~1
        in the DualVLN architecture, evaluated with \monitor and human-assisted recovery.
        Shading groups rows by architecture.
    \end{minipage}
    \vspace{-10pt} 
\end{table*}

%% file: figures/monitor-diagnostics-recognition.tex
\begin{figure*}[t]
    \centering
    \includegraphics[width=\textwidth]{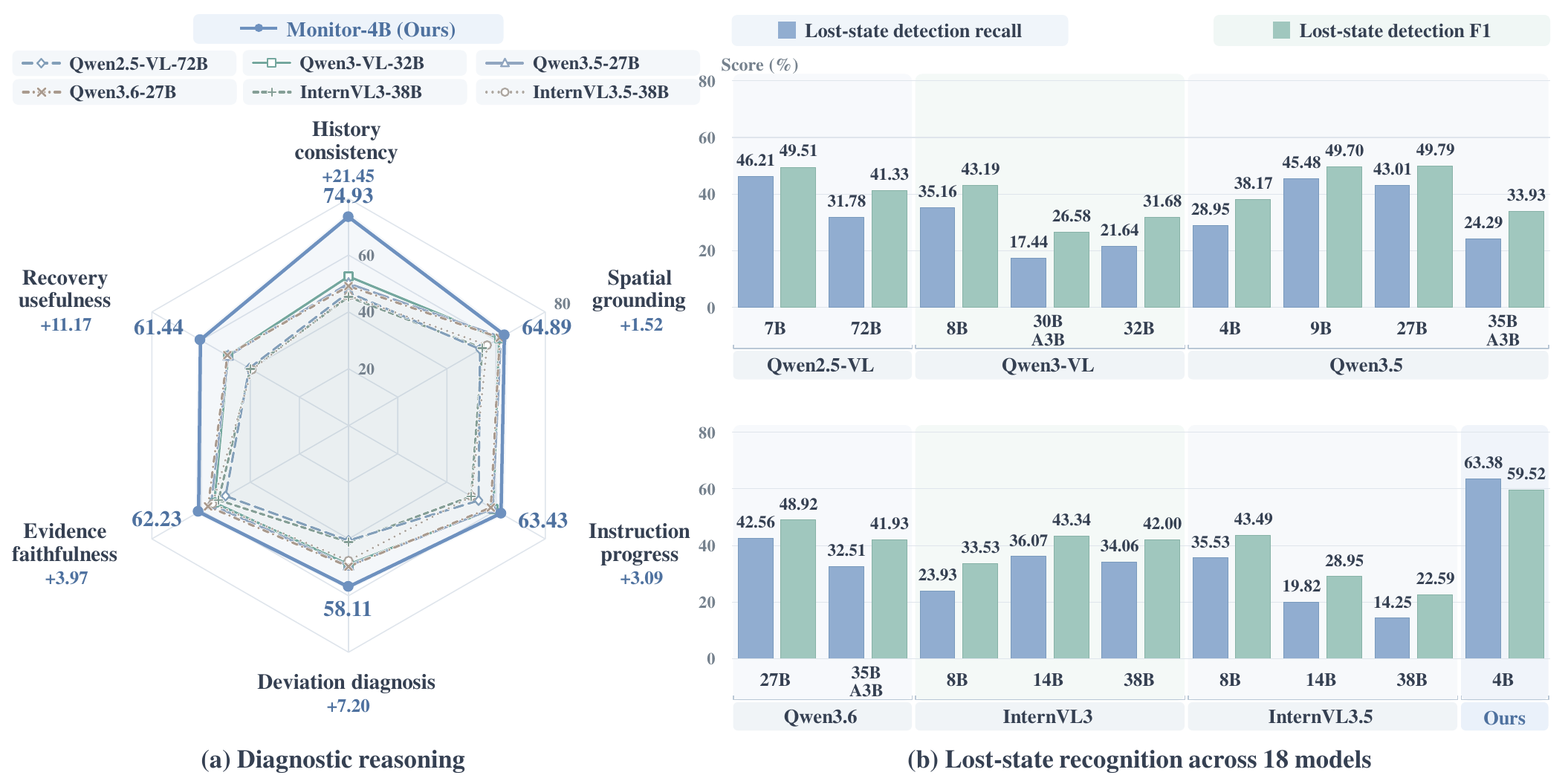}
    \caption{\textbf{Diagnostic reasoning and lost-state recognition.}
    (a) Six-dimensional diagnostic comparison of Monitor-4B and six selected baselines. Annotations report Monitor scores and gaps to the highest displayed baseline score in each dimension.
    (b) Lost-state detection recall and F1 (\%) for all 18 models on the 2,190 states in \benchmark, with \loststate as the positive class. Higher is better for both metrics.}
    \label{fig:monitor_radar}
    \vspace{-10pt} 
\end{figure*}

%% file: figures/dpo_qualitative_cases.tex
\begin{figure*}[!t]
    \centering
    \includegraphics[width=\textwidth]{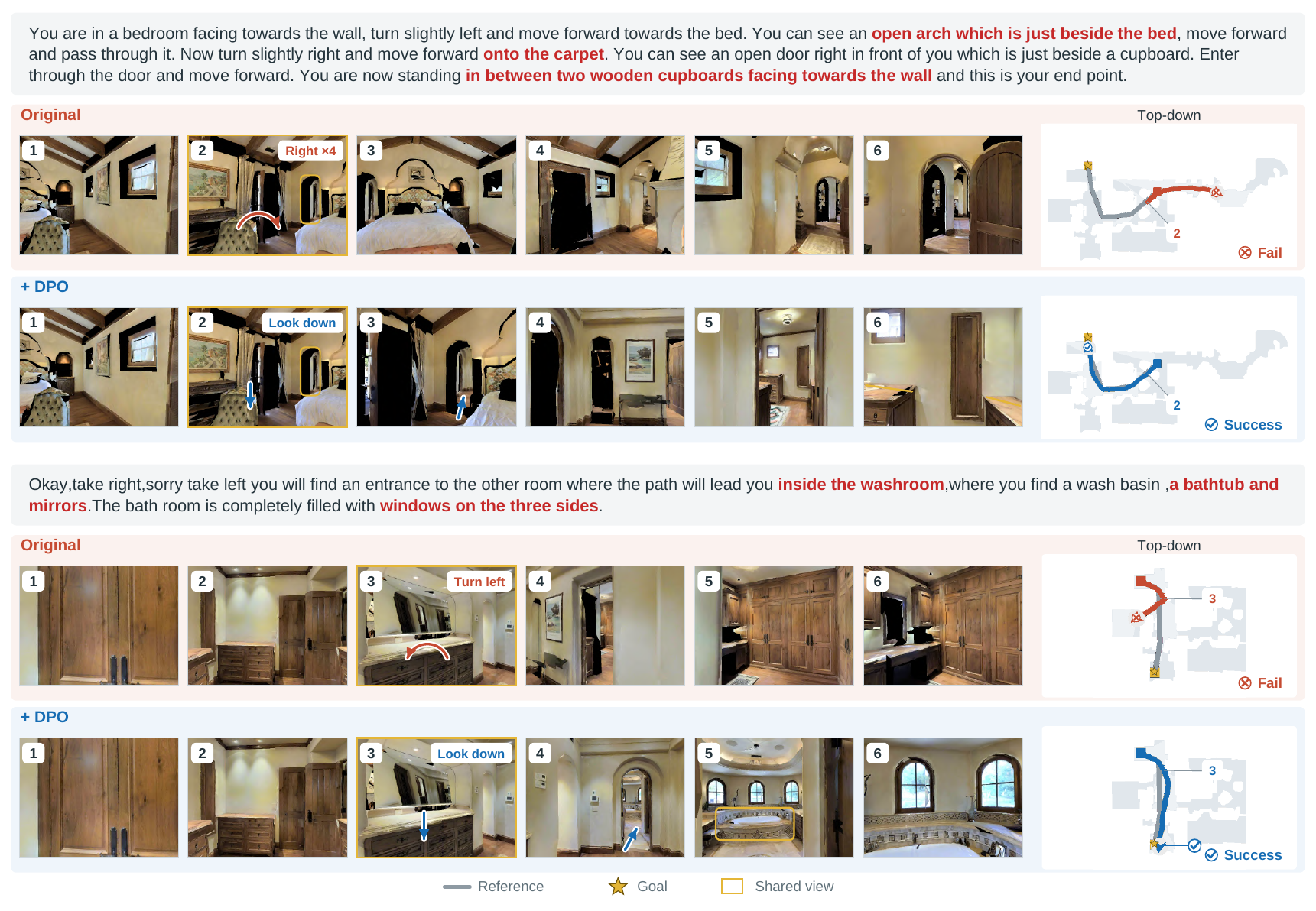}
    \caption{\textbf{Qualitative comparison of autonomous navigation before and after DPO.}
    Two selected RxR-CE val-unseen episodes are shown for the original System~2 policy (Original) and its DPO-trained counterpart (+DPO), with online monitoring and human assistance disabled.
    Each row presents six chronologically ordered observations and the final trajectory map; columns are not temporally aligned across rows.
    Gold borders mark shared observations; arrows indicate camera motions or navigation actions.}
    \label{fig:dpo_qualitative_cases}
    \vspace{-10pt} 
\end{figure*}

%% file: tables/ablation.tex
\begin{table}[!tb]
    \caption{\textbf{Component Ablation on RxR-CE Val-Unseen.}}
    \label{tab:component_ablation}
    \centering
    \footnotesize
    \setlength{\tabcolsep}{5.0pt}
    \renewcommand{\arraystretch}{1.10}

    \begin{NiceTabular}{lcccc}
        \CodeBefore
        \rowcolor{interncolor}{6}
        \Body
        \toprule
        \textbf{Configuration} & \textbf{NE} $\downarrow$ & \textbf{SR} $\uparrow$ & \textbf{OS} $\uparrow$ & \textbf{nDTW} $\uparrow$ \\
        \midrule
        System~2 & 5.77 & 52.88 & 65.30 & \uline{64.16} \\
        System~2 + SFT & 5.69 & 53.41 & 65.30 & \textbf{64.20} \\
        System~2 + DPO & 5.76 & 55.03 & 65.56 & 63.21 \\
        System~2 + \monitor & \textbf{4.53} & \uline{57.78} & \uline{68.71} & 57.25 \\
        System~2 + DPO + \monitor & \uline{4.67} & \textbf{60.21} & \textbf{70.29} & 57.13 \\
        \bottomrule
    \end{NiceTabular}
    \vspace{-10pt} 
\end{table}

%% file: sections/05_limitations_conclusion.tex
\section{Conclusion}
\label{sec:conclusion}

In this work, we presented \methodabbr, a continuous framework for deploying VLN agents in unseen environments.
It detects off-track execution, triggers human-assisted recovery for the ongoing task, and converts deployment failures into decision-level preference supervision for policy improvement.
Its asynchronous monitoring and recovery interface preserves the navigation model's native action-generation process, enabling integration across different navigation architectures.
Experiments on R2R-CE and RxR-CE show that human-assisted recovery improves task success across multiple navigation architectures, while autonomous ablations confirm that failure-driven preference learning improves subsequent navigation without online assistance.
Together, these results demonstrate the value of coupling selective intervention with failure-driven learning so that deployment failures can be recovered online and reused to improve future navigation.
\vspace{-5pt} 